\PassOptionsToPackage{most}{tcolorbox} 
\documentclass[11pt]{article}

\usepackage[final]{acl}

\usepackage{times}
\usepackage{latexsym}

\usepackage[utf8]{inputenc}
\usepackage{amsmath,amssymb}
\usepackage{pifont}
\usepackage{placeins} 
\usepackage{booktabs}
\usepackage{multirow}
\usepackage{lipsum}
\usepackage{tabularx}
\usepackage{subcaption}
\usepackage{arydshln}
\usepackage{adjustbox} 
\usepackage{makecell}
\usepackage{wrapfig}

\usepackage{siunitx}
\usepackage[T1]{fontenc}

\usepackage{inconsolata}
\usepackage{dblfloatfix}
\usepackage{graphicx}
\usepackage{enumitem}
\usepackage{cuted}
\usepackage{capt-of} 
\usepackage{graphicx}
\usepackage{dblfloatfix} 
\usepackage{caption}  
\usepackage{fontawesome5} 
\usepackage{hyperref}     
\usepackage{xcolor}       

\title{EvoFSM: Controllable Self-Evolution for Deep Research with Finite State Machines}

\author{
\textbf{Shuo Zhang\textsuperscript{1*}},
\textbf{Chaofa Yuan\textsuperscript{1*}},
\textbf{Ryan Guo\textsuperscript{1*}},
\textbf{Xiaomin Yu\textsuperscript{2}},
\textbf{Rui Xu\textsuperscript{3}},
\textbf{Zhangquan Chen\textsuperscript{4}},
\\
\textbf{Zinuo Li\textsuperscript{1}},
\textbf{Zhi Yang\textsuperscript{5}},
\textbf{Shuhao Guan\textsuperscript{6}},
\textbf{Zhenheng Tang\textsuperscript{7}},
\textbf{Sen Hu\textsuperscript{1,8}},
\textbf{Liwen Zhang\textsuperscript{5\dag}},
\\
\textbf{Ronghao Chen\textsuperscript{1,8\dag}},
\textbf{Huacan Wang\textsuperscript{1,9\dag}}
\\
\\
 \textsuperscript{1}QuantaAlpha,
 \textsuperscript{2}HKUST(GZ),
 \textsuperscript{3}FDU,
 \textsuperscript{4}THU,
 \textsuperscript{5}SUFE,
 \textsuperscript{6}UCD,
 \textsuperscript{7}HKUST,
 \textsuperscript{8}PKU, 
 \textsuperscript{9}UCAS, 
\\
\small {
    \textbf{\textsuperscript{*}These authors contributed equally to this work.}
}
\\
 \small{
   \textbf{\dag Correspondence:} 
   \href{mailto:zhang.liwen@shufe.edu.cn}{zhang.liwen@shufe.edu.cn},
   \href{mailto:chenronghao@alumni.pku.edu.cn}{chenronghao@alumni.pku.edu.cn},
   \href{mailto:wanghuacan17@mails.ucas.ac.cn}{wanghuacan17@mails.ucas.ac.cn}
 }
}

\begin{document}
\maketitle

\begin{center}
    \vspace{-15pt}
    \large
    \faGithub\hspace{6pt}\href{https://github.com/QuantaAlpha/EvoFSM}{\texttt{\color{black}https://github.com/QuantaAlpha/EvoFSM}}
\end{center}

\vspace{10pt} 

\begin{abstract}
While LLM-based agents have shown promise for deep research, most existing approaches rely on fixed workflows that struggle to adapt to real-world, open-ended queries. Recent work therefore explores self-evolution by allowing agents to rewrite their own code or prompts to improve problem-solving ability, but unconstrained optimization often triggers instability, hallucinations, and instruction drift. We propose \textbf{EvoFSM}, a structured self-evolving framework that achieves both adaptability and control by evolving an explicit Finite State Machine (FSM) instead of relying on free-form rewriting. EvoFSM decouples the optimization space into macroscopic \textit{Flow} (state-transition logic) and microscopic \textit{Skill} (state-specific behaviors), enabling targeted improvements under clear behavioral boundaries. Guided by a critic mechanism, EvoFSM refines the FSM through a small set of constrained operations, and further incorporates a self-evolving memory that distills successful trajectories as reusable priors and failure patterns as constraints for future queries. Extensive evaluations on five multi-hop QA benchmarks demonstrate the effectiveness of EvoFSM. In particular, EvoFSM reaches 58.0\% accuracy on the DeepSearch benchmark. Additional results on interactive decision-making tasks further validate its generalization.

\end{abstract}

\begin{figure}[h]
    \centering
    \includegraphics[width=0.75\textwidth]{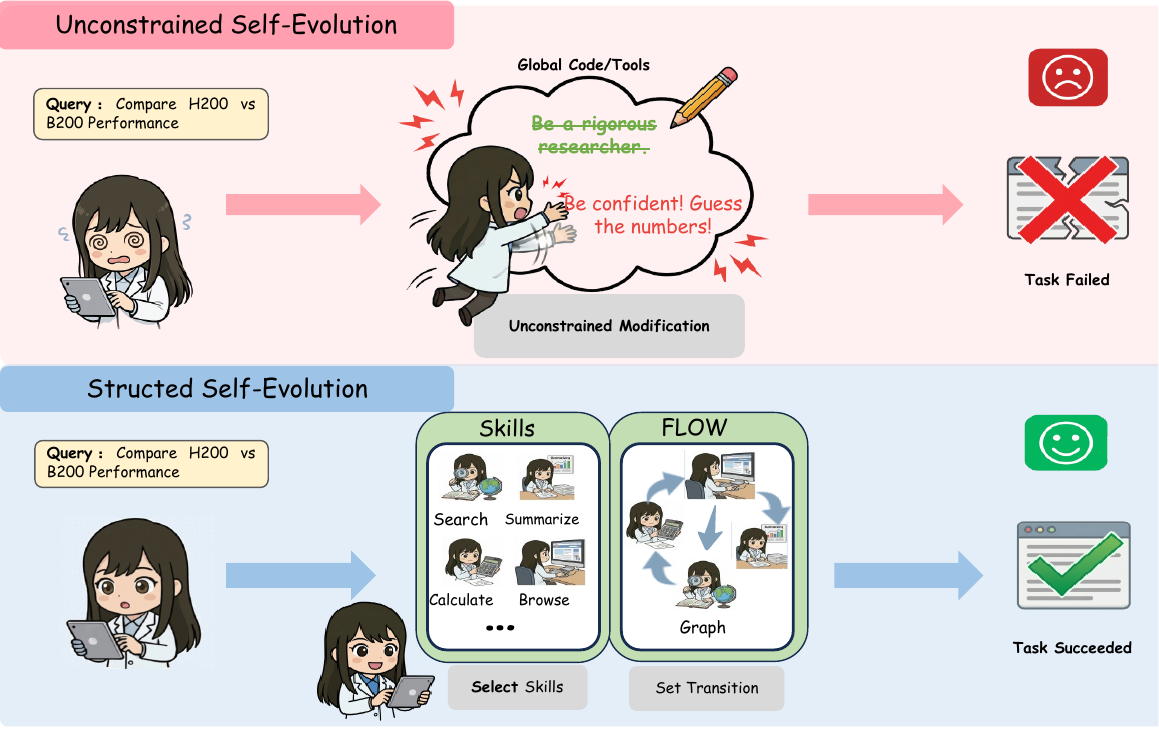}
    \caption{Comparison of unconstrained self-evolution and our structured self-evolution.}
    \label{fig:intro} 
\end{figure}

\newpage

\section{Introduction}

Large Language Model (LLM) based agents have demonstrated remarkable progress in automating Deep Research tasks~\citep{li2025searcho1, jin2025searchr1, li2025webthinker}. However, dominant approaches still rely heavily on pre-defined, static procedural paradigms, typically characterized by fixed, iterative ``toolcall-generation-reflection'' pipelines~\citep{wei2025webagentr1, zhao2025parallelsearch}. The inherent rigidity of these structures struggles to adapt to the dynamic query paths required by real-world, open-ended research problems, significantly constraining system flexibility and generalization capabilities.

To address the limitations of static workflows, the research community has pivoted towards ``Self-Evolution'' mechanisms~\citep{deppen2024live, wang2024stella, li2024morphagent}, aiming to empower agents to dynamically adjust strategies based on task feedback. However, existing self-evolution methods often employ ``unconstrained rewriting,'' where a Meta-Agent is granted the freedom to rewrite the entire system prompt or tools of worker agents~\citep{schmidhuber2024huxley}. This unconstrained approach poses significant stability challenges, as agents may suffer from core instruction drift, hallucination, or the corruption of robust functional modules during self-modification, potentially causing the system to deteriorate rather than improve~\citep{yuan2024misevolve}, as shown in Figure~\ref{fig:intro}.

In human organizational collaboration, efficiency gains typically stem from the joint optimization of two dimensions: workflow planning and specific execution skills. Drawing inspiration from this, we posit that agent evolution should not be a chaotic global rewriting process, but rather a form of structured and controllable optimization. We explicitly decouple the optimization space into two orthogonal dimensions: \textit{Flow} (i.e., macroscopic transition logic) and \textit{Skill} (i.e., microscopic node-specific capabilities). This decomposition transforms the self-evolving process from chaotic modification into a targeted, structured self-evolution mechanism.

In this paper, we propose \textbf{EvoFSM}, a structured self-evolving framework designed for deep research and general enough to be applied across diverse tasks and domains. Specifically, EvoFSM first models the complex retrieval-reasoning process as an explicit Finite State Machine (FSM) \citep{wu2024stateflow}. By decomposing uncertain, long-horizon tasks into a state graph with clear transition logic, we establish deterministic behavioral boundaries that guarantee foundational stability. Second, to mitigate the uncontrollability of evolution, EvoFSM employs a ``Structured Self-Evolution'' mechanism. Rather than allowing free-form rewriting, we restrict the system to modifying the FSM topology only via a set of atomic operations guided by a critic mechanism. This targeted adjustment ensures the system flexibly adapts to new tasks without compromising functional integrity. Finally, to accumulate experience across tasks and enable continual improvement, we introduce a self-evolving memory mechanism that stores successful strategies as priors and failure patterns as constraints for future queries.

By integrating these mechanisms, EvoFSM establishes a structured, self-evolving system. In summary, our main contributions are as follows:

\begin{itemize}
    \item We introduce EvoFSM, a structured self-evolution framework that models task solving as an FSM, enabling controllable evolution through explicit states and transition logic rather than unconstrained rewriting.
    \item We introduce a self-evolving memory mechanism that accumulates experience across tasks and uses it to guide future evolution, enabling continual improvement beyond per-task optimization.
    \item Extensive evaluations on five multi-hop QA benchmarks demonstrate the effectiveness of EvoFSM, and results on two interactive decision-making datasets further support its generality across settings.
\end{itemize}

\section{Related Work}

\subsection{Deep Research Agents}
To tackle open-ended and complex information-seeking tasks, recent work has proposed a range of deep research agents. Search-o1 \citep{li2025searcho1} integrates agentic RAG with a Reason-in-Documents module to support on-demand knowledge acquisition during long-horizon reasoning. Search-R1 \citep{jin2025searchr1} further trains LLMs via reinforcement learning to generate search queries and interact with search engines over multiple turns. For richer web interaction, WebThinker \citep{li2025webthinker} enables autonomous browsing and report drafting, while WebAgent-R1 \citep{wei2025webagentr1} uses end-to-end reinforcement learning for decision-making in dynamic web environments. RepoMaster \citep{wang2025repomaster} extends agentic exploration to software engineering through repository structure graphs. Efficiency-oriented methods such as ReSearch \citep{sun2024research} and ZeroSearch \citep{xu2024zerosearch} improve ranking or reduce supervision. In contrast, EvoFSM uses an explicit FSM to organize deep research, decomposing the process into specialized states with clear transition logic. Crucially, it can adapt the FSM to each query by evolving the workflow based on task feedback and prior experience.

\subsection{Self-Evolving Agents}
Recent work has explored self-evolving agents that improve during deployment \citep{shinn2023reflexion, madaan2023selfrefine, yang2023llmasoptimizers}. In code and tool domains, LIVE-SWE-AGENT \citep{deppen2024live} and STELLA \citep{wang2024stella} allow agents to synthesize tools or update their scaffolding on the fly, while ShieldLearner \citep{ni2025shieldlearner} learns reusable defenses against jailbreaks. Self-evolution has also been extended to multi-agent settings, e.g., MorphAgent \citep{li2024morphagent} adapts agent profiles and CoMAS \citep{wu2025comas} drives decentralized co-evolution via interaction rewards. More radical proposals such as the Huxley-Gödel Machine \citep{schmidhuber2024huxley} pursue open-ended self-improvement through recursive modifications, and SE-Agent \citep{lin2025seagent} optimizes reasoning trajectories through multi-step interaction. Despite these advances, many approaches rely on unconstrained prompt rewriting or global architecture search with limited structural guardrails, which can lead to instability and failed evolution \citep{yuan2024misevolve}. In contrast, our EvoFSM introduces explicit structural guardrails and experience-guided evolution to keep self-improvement stable. It adapts its strategy to each query while preserving core functionality, mitigating the instruction drift and instability commonly observed in unconstrained self-evolution.

\begin{figure*}[t]
    \centering
    \includegraphics[width=\textwidth]{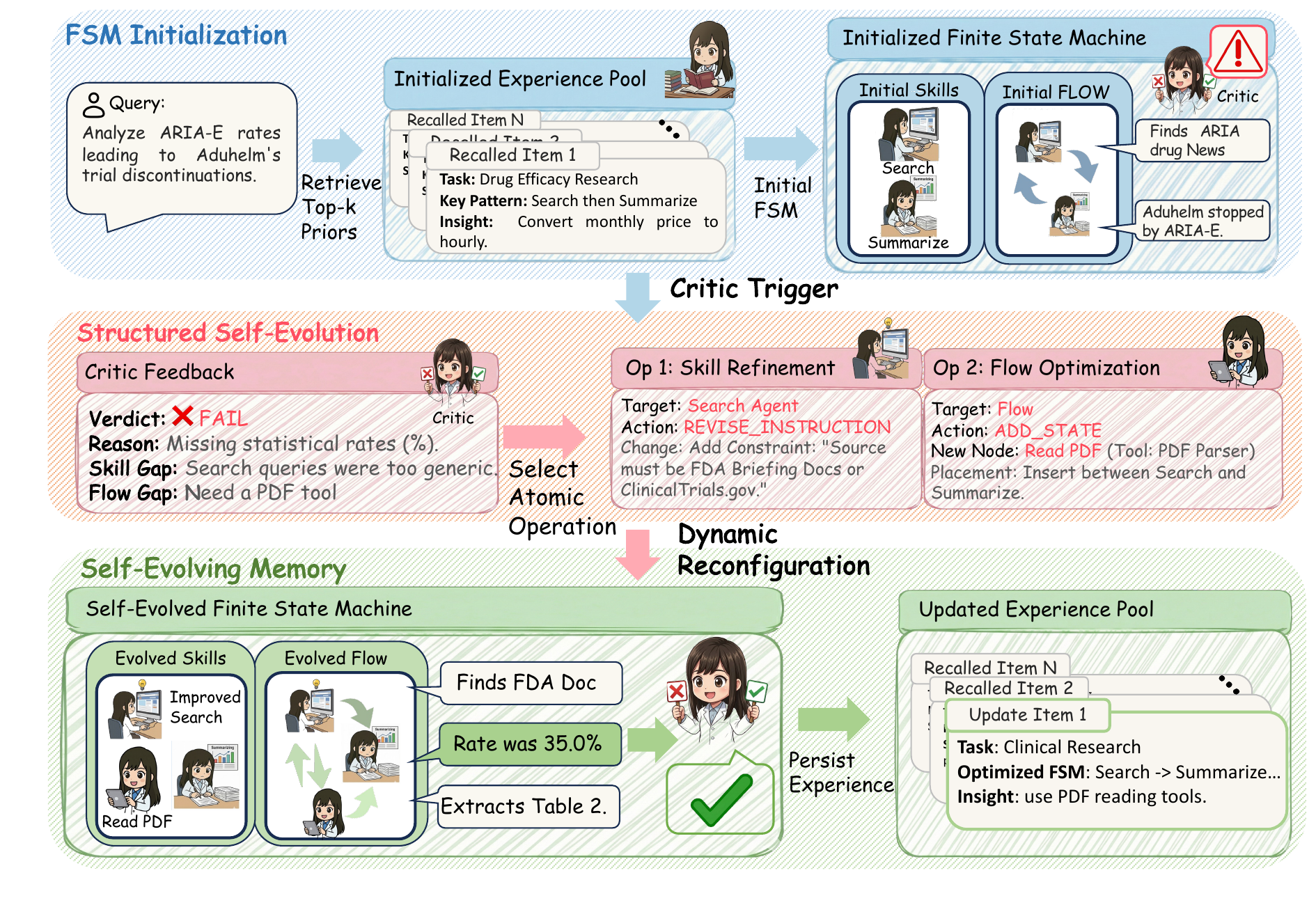}
    \caption{Overview of the EvoFSM framework. Our approach consists of three core components: 
    (1) FSM Initialization, which formalizes the research process as a dynamic finite state machine initialized from prior experiences; 
    (2) Structured Self-Evolution, which employs atomic operations to precisely optimize both the system's skill operators ($\mathcal{O}_{skill}$) and flow operators ($\mathcal{O}_{flow}$) based on critic feedback; and 
    (3) Self-Evolving Memory Mechanism, which distills successful and failure trajectories into an experience pool to facilitate continuous learning and warm-starting for future tasks.}
    \label{fig:framework}
\end{figure*}

\section{Methodology}

In this section, we present the EvoFSM framework, as illustrated in Figure~\ref{fig:framework}. 
EvoFSM comprises three core components: FSM Initialization, Structured Self-Evolution, and the Self-Evolving Memory Mechanism. 
We detail each of these components in the following subsections.

\subsection{FSM Initialization}
\label{sec:fsm_modeling}

While recent works have explored self-evolution to improve adaptability, these approaches often rely on an unconstrained rewriting mechanism~\citep{yuan2024misevolve}, which frequently leads to hallucination and instruction drift. To address the uncontrollable evolution, we model the deep research process as a deterministic yet dynamic FSM. 
Fundamentally, the FSMs serve as a directed behavioral graph, where nodes represent specialized action phases (e.g., Search, Browse) and edges define the logical transitions between them based on runtime context. This graph-based structure grounds the agent's behaviors, serving as a robust backbone that enables the precise, targeted self-evolving optimization described in Section~\ref{sec: structured_evolution}. We formalize this graph-based system as a tuple $\mathcal{M} = \langle \mathcal{S}, \mathcal{T}, \mathcal{I}, \mathcal{C} \rangle$, structured to align with the dual optimization dimensions of \emph{Flow} and \emph{Skill}:

\begin{itemize}

    \item $\mathcal{S}$ represents the extensible set of cognitive states (nodes), initially comprising fundamental capabilities such as \textit{Problem Decomposition}, \textit{Search}, and \textit{Browsing}.
    
    \item $\mathcal{T}: \mathcal{S} \times \mathcal{H} \rightarrow \mathcal{S}$ constitutes the macroscopic Flow logic. It functions as a dynamic router that determines the next state $s_{t+1}$ based on the current state $s_t$ and runtime context $\mathcal{H}$, thereby shaping the sequential execution path of the collaboration.
    
    \item $\mathcal{I}$ represents the set of node-specific system prompts, defining the microscopic Skill dimension. Each instruction $i \in \mathcal{I}$ specifies the operational guidelines and expertise for its corresponding agent.

   \item $\mathcal{C}$ denotes the Critic Mechanism, a supervisory module that evaluates the final output against the user query. Unlike passive evaluation metrics, $\mathcal{C}$ serves as the active trigger for the self-evolving process, identifying specific failure modes (e.g., lack of quantitative evidence or logical inconsistencies)

\end{itemize}

Based on the formal definitions above, the FSM initialization proceeds as follows:

When the system receives a user query $q$, the framework first retrieves the top-$k$ relevant historical strategies from the Experience Pool $\mathcal{E}$, which accumulates the system's historical successful and failed trajectories from previously executed tasks and detailed in Section~\ref{sec:evolution_memory}, to initialize the FSM configuration, denoted as $\mathcal{M}_{init}$. During the execution phase, the agents first perform the node-specific actions (Skill $\mathcal{I}$) defined by the current state. Since the topology of $\mathcal{M}_{init}$ remains static during this phase, the agent may encounter capability gaps or become trapped in unproductive loops when confronting novel, complex scenarios, especially in deep research scenarios. Consequently, following the execution phase, the critic mechanism $\mathcal{C}$ validates the system's output against the query requirements, where detected failures (e.g., hallucination or incomplete reasoning) serve as trigger signals to activate the Structured Self-Evolution (detailed in Section~\ref{sec: structured_evolution}) to dynamically reconfigure the FSM structure.

\subsection{Structured Self-Evolution}
\label{sec: structured_evolution}

Departing from unconstrained modification, EvoFSM constrains the evolution process to a structured mechanism grounded in discrete atomic operations. By leveraging the explicit components of the tuple $\mathcal{M}$, we decouple the optimization space into two dimensions. First, \textit{Flow Operators}, denoted as $\mathcal{O}_{flow}$, reconfigure the collaboration topology governed by the transition function $\mathcal{T}$. Second, \textit{Skill Refinement}, denoted as $\mathcal{O}_{skill}$, enhances the individual expertise of each agent governed by the instruction set $\mathcal{I}$.

The framework evolves the FSM exclusively through a strictly defined set of Atomic Operations. Let $\mathcal{M}_t$ represent the FSM structure at iteration $t$. The evolution process is formulated as $\mathcal{M}_{t+1} = \mathcal{M}_t \oplus \textit{op}$, where the operation $\textit{op}$ is chosen from one of two operator sets, corresponding to Flow- and Skill-level evolution.

\paragraph{Flow Operators $\mathcal{O}_{flow}$}
These operations reconfigure the collaboration topology by editing the state set and/or the transition logic, without changing any node-specific instructions.
\begin{itemize}
    \item \texttt{ADD\_STATE} inserts an intermediate state (e.g., a verification state) to bridge a workflow gap.
    \item \texttt{DELETE\_STATE} removes redundant or low-utility states to streamline execution.
    \item \texttt{MODIFY\_TRANSITION} adjusts transition conditions (e.g., revisiting Search when the retrieved evidence is insufficient).
\end{itemize}

\paragraph{Skill Operators $\mathcal{O}_{skill}$}
These operations refine node-specific instructions while keeping the global topology unchanged.
\begin{itemize}
    \item \texttt{REVISE\_INSTRUCTION} updates the guidelines or constraints for a single state (e.g., instructing the browsing state to prioritize PDFs over news sources to improve evidence quality).
\end{itemize}

By enforcing these atomic operations, EvoFSM ensures that any modification remains local, interpretable, and reversible.

\subsection{Self-Evolving Memory}
\label{sec:evolution_memory}

While the structured evolution framework empowers agents to dynamically adapt to specific queries, treating each task as an isolated optimization problem prevents the system from accumulating valuable experience. To address this limitation and enable continuous improvement, EvoFSM integrates a self-evolving memory mechanism that functions as a repository of successful strategies.

The system maintains an experience pool $\mathcal{E}$ to persist high-quality self-evolving patterns. Upon the completion of a task, a specific reflection agent analyzes the execution trajectory to determine if the task was resolved successfully. In such cases, the final optimized FSM configuration and the sequence of atomic operations are distilled into a strategy record $r$. This record encapsulates the query embedding $q_{\text{embedding}}$, the optimized model structure $\mathcal{M}_{optimized}$, and the rationale for the changes.

When the system receives a new query $q_{new}$, the module retrieves the top-$k$ historical records from the experience pool $\mathcal{E}$, where $\mathcal{E} = \mathcal{E}^{+} \cup \mathcal{E}^{-}$. In this context, successful strategies ($\mathcal{E}^{+}$) serve as initialization priors, allowing the agent to inherit effective execution trajectories (e.g., initializing with a search-verify loop for medical diagnoses as shown in Figure~\ref{fig:framework}). Conversely, retrieved failure patterns ($\mathcal{E}^{-}$) serve as negative constraints, actively warning the transition logic against specific transition paths or tool usages that previously led to dead ends in similar contexts. This mechanism enables the system to exhibit progressively stronger performance by building upon past experiences.

\begin{table*}[t!]
\centering
\begin{tabular*}{\textwidth}{@{} l l @{\extracolsep{\fill}} c c c c c @{}}
\toprule
\multicolumn{2}{c}{\multirow{2}{*}{\textbf{Model \& Framework}}} &
\multicolumn{5}{c}{\textbf{Multi-hop QA Benchmarks}} \\
\cmidrule(lr){3-7}
& & \textbf{HotpotQA} & \textbf{2WIKI} & \textbf{MuSiQue} & \textbf{Bamboogle} & \textbf{Deepsearch} \\
\midrule
\multirow{4}{*}{\makecell{GPT-4o}}
  & Standard RAG & 68.2 & 54.2 & 42.8 & 80.0 & 18.0 \\
  & Agentic RAG  & 76.6 & 71.4 & 41.6 & \underline{84.0} & 33.0 \\
  & Search-o1    & \underline{77.8} & \underline{85.6} & \underline{48.2} & \textbf{87.2} & \underline{35.0} \\
  & EvoFSM  &
    \textbf{80.2} &
    \textbf{88.8} &
    \textbf{50.4} &
    82.4 &
    \textbf{45.0} \\
\hdashline
\multirow{4}{*}{\makecell{Claude-4}}
  & Standard RAG & 65.4 & 30.8 & 33.4 & 75.2 & 17.0 \\
  & Agentic RAG  & 80.6 & 90.0 & \underline{51.0} & \underline{88.8} & \underline{53.0} \\
  & Search-o1    & \underline{81.2} & \underline{91.6} & \underline{51.0} & 87.2 & 47.0 \\
  & EvoFSM &
    \textbf{82.2} &
    \textbf{91.8} &
    \textbf{57.6} &
    \textbf{91.2} &
    \textbf{58.0} \\
\hdashline
\multirow{4}{*}{\makecell{Llama3-70B}}
  & Standard RAG & 72.2 & 57.0 & 40.4 & \textbf{84.0} & 16.0 \\
  & Agentic RAG  & 61.6 & 74.4 & 38.0 & 67.2 & 23.0 \\
  & Search-o1    & \underline{76.2} & \textbf{85.2} & \underline{43.0} & \underline{83.2} & \underline{24.0} \\
  & EvoFSM &
    \textbf{76.6} &
    \underline{75.6} &
    \textbf{46.4} &
    80.4 &
    \textbf{28.0} \\
\hdashline
\multirow{4}{*}{\makecell{DeepSeek-v3}}
  & Standard RAG & 72.6 & 51.8 & 41.0 & 82.4 & 30.0 \\
  & Agentic RAG  & 65.2 & 71.2 & 42.0 & 66.4 & 31.0 \\
  & Search-o1    & \underline{74.6} & \underline{85.0} & \underline{46.6} & \underline{84.0} & \underline{43.0} \\
  & EvoFSM &
    \textbf{80.4} &
    \textbf{88.8} &
    \textbf{\underline{54.2}} &
    \textbf{89.6} &
    \textbf{51.0} \\
\hdashline
\multirow{4}{*}{\makecell{Qwen3-32B}}
  & Standard RAG & 63.2 & 41.6 & 33.8 & 74.4 & 20.0 \\
  & Agentic RAG  & 64.6 & 69.6 & 34.8 & 68.8 & 19.0 \\
  & Search-o1    & \underline{67.0} & \underline{76.2} & \underline{37.2} & \textbf{84.0} & \underline{27.0} \\
  & EvoFSM &
    \textbf{77.8} &
    \textbf{83.6} &
    \textbf{43.8} &
    \underline{81.6} &
    \textbf{32.0} \\
\bottomrule
\end{tabular*}
\caption{Accuracy (\%) of different retrieval frameworks for each backbone model on five challenging multi-hop QA benchmarks. Best results are in \textbf{bold} and second-best results are \underline{underlined}.}
\label{tab:retrieval_framework_comparison}
\end{table*}

\section{Experiments}

\subsection{Experimental Settings}
\paragraph{Datasets and Metrics.} We evaluate our EvoFSM on five multi-hop question answering (QA) benchmarks that require aggregating evidence across multiple documents. Specifically, we consider the following benchmarks: HotpotQA \citep{yang2018hotpotqadatasetdiverseexplainable}, the first large-scale dataset requiring reasoning over multiple Wikipedia paragraphs; 2WikiMultihopQA (2WIKI) \citep{ho-etal-2020-constructing}, which provides multi-hop QA with explicit reasoning paths; MuSiQue \citep{trivedi-etal-2022-musique}, which synthesizes 2–4 hop questions by composing evidence from five single-hop datasets; Bamboogle \citep{press-etal-2023-measuring}, a collection of compositional questions that search engines often answer incorrectly; and xbench-DeepSearch (Deepsearch) \citep{chen2025xbench}, part of the xbench AGI-Aligned series, which provides a Chinese-context question pool for multi-hop QA and deep search.

\paragraph{Baselines.} We compare EvoFSM against three baseline methods. Standard RAG performs a single web-search round and appends the top$-10$ retrieved documents to the prompt as additional context. Agentic RAG\footnote{We follow the implementation details in \cite{li2025searcho1}.} introduces an iterative retrieve–reason loop, allowing the model to autonomously issue additional retrieval requests when the current evidence is insufficient. Search-o1 \cite{li2025searcho1} further strengthens the agentic RAG pipeline by adding a \emph{Reason-in-Documents} stage, which distills retrieved passages into concise, relevant evidence while preserving the intermediate reasoning process. In addition, to demonstrate the generality of our method, we conduct experiments with a diverse set of LLMs, including proprietary models (GPT-4o\citep{gpt4o} and Claude-4\citep{claude4}) and open-weight models (Llama-3-70B\citep{grattafiori2024llama3herdmodels}, DeepSeek-V3\citep{liu2024deepseek}, and Qwen3-32B\cite{yang2025qwen3technicalreport}).

\paragraph{Implementation Details.} Our system builds on AutoGen \citep{wu2023autogenenablingnextgenllm} for multi-agent orchestration, enabling reliable inter-agent communication and stateful execution. We further provide a tool library, including the Serper\footnote{\url{https://serper.dev}} API for retrieval and the Jina Reader\footnote{\url{https://jina.ai}}  API for web-page content extraction. To ensure fair comparison of structural efficiency and avoid overfitting, we limit the optimization loop to 3 iterations. We also cap the number of states at 10 to prevent unbounded growth, while still permitting dynamic refinements to state definitions and transitions when necessary.

\subsection{Main Results}

Table~\ref{tab:retrieval_framework_comparison} presents the performance comparison of different frameworks across five diverse benchmarks. From these results, we draw three observations.

First, iterative retrieval-and-reasoning frameworks consistently outperform Standard RAG across all benchmarks. For instance, under GPT-4o, Agentic RAG delivers a 15.0\% absolute gain on DeepSearch over the single-shot retrieval baseline. This gap reflects a fundamental limitation of one-pass retrieval: the system must surface all necessary evidence in a single query, with limited opportunity to recover from missing or off-target context. In contrast, iterative methods can progressively refine queries and validate intermediate hypotheses against newly retrieved evidence. EvoFSM retains this iterative capability, but makes it more reliable by explicitly structuring the loop into well-defined phases and transitions, reducing wasted iterations and stabilizing evidence accumulation.

Second, the relative comparison among frameworks is largely stable across the evaluated LLMs. In other words, the improvements brought by EvoFSM are not tied to a single backbone or a narrow operating regime; instead, the same design advantages tend to persist when the underlying model changes. This robustness suggests that the performance gains primarily come from the workflow and optimization mechanisms, rather than from model-specific quirks.

Finally, EvoFSM delivers the strongest overall results and a consistent margin over both Agentic RAG and Search-o1. In particular, EvoFSM consistently outperforms Search-o1 across all five LLMs on DeepSearch. For example, it improves by 11.0\% with Claude-4 and 10.0\% with GPT-4o, and still yields a 4.0\% gain with Llama-3-70B. We attribute this additional margin to structured evolution, which makes the iterative retrieve–reason process experience-guided and adaptively evolvable. EvoFSM initializes each new query with relevant historical strategies and then refines its workflow to the specific instance, enabling more robust evidence acquisition and stronger multi-hop reasoning under the same tool budget.

\subsection{Ablation Study}
To disentangle the contribution of each component in EvoFSM, we perform ablation studies with DeepSeek-v3 as the backbone on five benchmarks. Specifically, we isolate the effects of (i) the structured self-evolution mechanism with memory and (ii) the explicit FSM topology. The results are summarized in Table~\ref{tab:ablation_study}.


\begin{table*}[t!]
\centering
\renewcommand{\arraystretch}{1.3} 

\resizebox{\textwidth}{!}{%
    \begin{tabular}{l c c c c c}
    \toprule
    \multirow{2}{*}{\textbf{Method}} & \multicolumn{5}{c}{\textbf{Multi-hop QA Benchmarks}} \\
    \cmidrule(lr){2-6}
    & \textbf{HotpotQA} & \textbf{2WIKI} & \textbf{MuSiQue} & \textbf{Bamboogle} & \textbf{DeepSearch} \\
    \midrule

    \textbf{EvoFSM} (Full Model) & 
    \textbf{80.4} & \textbf{88.8} & \textbf{54.2} & \textbf{89.6} & \textbf{51.0} \\

    \quad \textbf{w/o Structured Evolution} (Static FSM) & 
    77.8 \small{(-2.6)} & 82.4 \small{(-6.4)} & 50.8 \small{(-3.4)}  & 85.6 \small{(-4.0)} & 36.0 \small{(-15.0)} \\

    \quad \textbf{w/o FSM Topology} (Unstructured Evolution) & 
    77.0 \small{(-3.4)} & 83.8 \small{(-5.0)} & 50.2 \small{(-4.0)} & 86.4 \small{(-3.2)} & 42.0 \small{(-9.0)} \\

    \quad \textbf{w/o All} (Standard ReAct) & 
    76.6 \small{(-3.8)} & 84.0 \small{(-4.8)} & 48.4 \small{(-5.8)} & 82.4 \small{(-7.2)} & 34.0 \small{(-17.0)} \\

    \bottomrule
    \end{tabular}%
} 

\caption{Ablation Study of EvoFSM framework. We systematically remove the structured self-evolution mechanism and the FSM topology to evaluate their contributions.}
\label{tab:ablation_study}
\end{table*}

\paragraph{Impact of Structured Self-Evolution.}
We first evaluate the role of the evolution mechanism by removing the optimization loop and memory retrieval, denoted as \textit{w/o Structured Evolution (Static FSM)}. This variant uses a fixed FSM initialized with a manually designed workflow. Specifically, it contains three predefined states—\textit{Search}, \textit{Browsing}, and \textit{Analysis}—together with hand-written instructions and deterministic transition rules. The transition logic routes the state from \textit{Search} to \textit{Browsing} after retrieving candidate sources, proceeds to \textit{Analysis} once sufficient evidence is collected, and otherwise falls back to \textit{Search} to issue refined queries. Since both the state set and the transition conditions are fixed, the workflow cannot be reconfigured during execution. Both the state set and the transition logic remain unchanged throughout execution.
The results reveal a substantial performance drop across all datasets, most notably on the DeepSearch benchmark, where accuracy plummets by 15.0\% (from 51.0\% to 36.0\%). This sharp decline underscores the critical limitation of static workflows: without the ability to dynamically apply atomic operations ($\mathcal{O}_{flow}$ and $\mathcal{O}_{skill}$) or recall successful priors from memory, the agent cannot adapt to unforeseen bottlenecks in complex queries, leading to repeated strategic failures such as insufficient verification or data extraction errors.
\paragraph{Necessity of FSM Topology.}
Next, we investigate the importance of the graph-based control structure by removing the explicit FSM constraints while retaining the self-evolution capability, denoted as \textit{w/o FSM Topology (Unstructured Evolution)}. In this setting, the system allows for free-form prompt rewriting but lacks the deterministic guidance of a state transition graph. We observe a performance degradation of 9.0\% on DeepSearch. While prompt evolution provides some adaptability, the absence of a rigid structural backbone leads to unstable behavior. Without the explicit \textit{Flow} logic governed by the FSM, agents are prone to redundant looping or losing track of the research objective, confirming that structural boundaries are essential for keeping self-evolution targeted and stable.
\paragraph{Synergy of Structure and Evolution.}
Finally, the \textit{w/o All (Standard ReAct)} baseline, which strips away both the FSM structure and the self-evolving mechanism, yields the lowest performance across the board, with a 17.0\% drop on DeepSearch. Comparing this baseline with the single-component ablations reveals a compounding effect: combining the FSM structure with self-evolution yields larger gains than either component alone. The FSM provides the necessary stability for long-horizon reasoning, while structured self-evolution ensures that this stability does not degenerate into rigidity. Together, they enable EvoFSM to achieve both high precision and robust adaptability.

\subsection{Further Analysis}

\begin{figure}[t]
    \centering
    \includegraphics[width=0.8\linewidth]{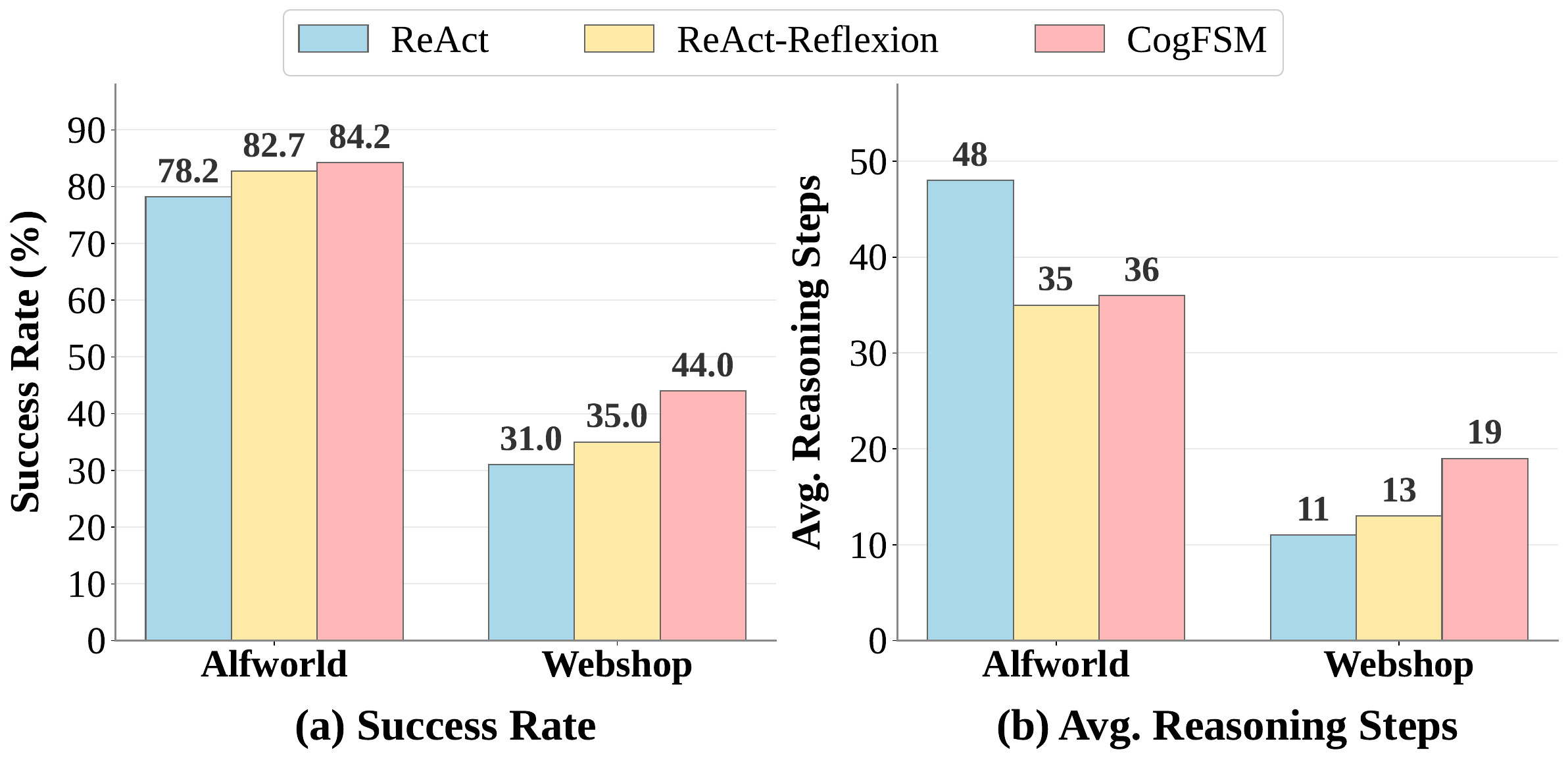} 
    \caption{Transferability study on the ALFWorld and WebShop benchmarks. We compare the success rate (a) and average reasoning steps (b) of each method.}
\label{fig:transfer_study}
\end{figure}

\paragraph{Cross-Domain Generalization Capabilities.}
\label{sec:transferability}
To assess the versatility of EvoFSM beyond deep research, we evaluate it on two interactive decision-making benchmarks, ALFWorld \citep{shridhar2021alfworld} and WebShop \citep{yao2023webshop}. ALFWorld is a text-based household environment containing 134 tasks spanning six types, such as object manipulation and navigation. The agent interacts through natural-language actions (e.g., moving to locations, picking up objects) and receives textual observations as feedback. WebShop simulates an online shopping platform with 1.18M real-world products and 12k human-written instructions, where the agent must satisfy user requirements by issuing search queries and selecting products through multi-step interactions.

As shown in Figure~\ref{fig:transfer_study}(a), EvoFSM consistently outperforms both ReAct \citep{yao2023react} and Reflexion \citep{shinn2023reflexion}, with particularly strong gains on WebShop. These results suggest that the structured FSM execution helps mitigate degenerate long-horizon behaviors such as redundant looping, and that the benefits of structured self-evolution transfer from deep research to general interactive tasks. In terms of efficiency, Figure~\ref{fig:transfer_study}(b) shows that EvoFSM uses slightly more reasoning steps than ReAct. This is expected, since the FSM explicitly allocates steps for validation and disciplined state transitions to improve robustness, trading a modest increase in steps for higher task success rates.




\paragraph{Impact of Optimization Iterations.}
\label{sec:hyperparameter}

\begin{wrapfigure}{r}{0.52\linewidth} 
    \centering
    \includegraphics[width=1\linewidth]{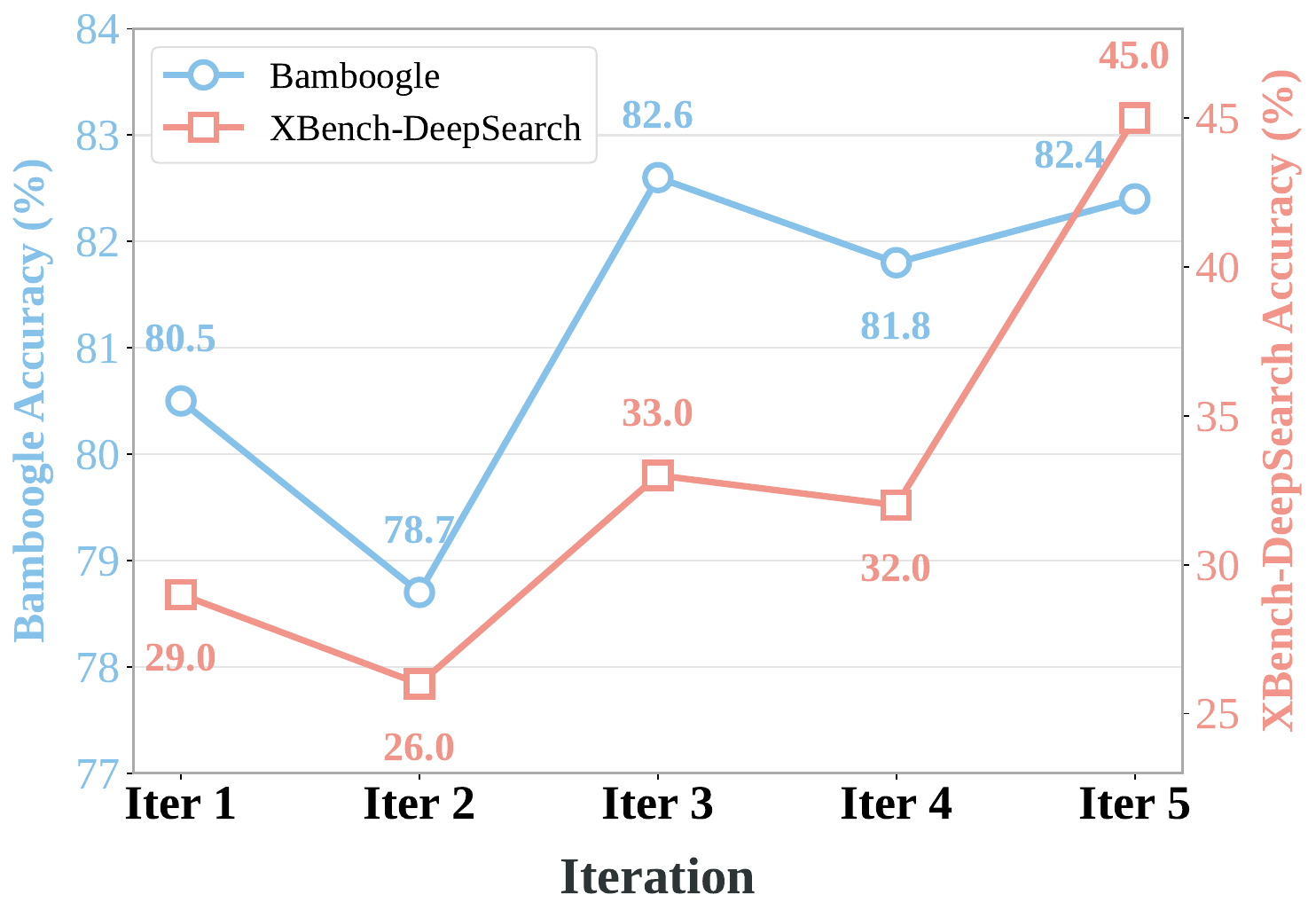}
    \caption{Effect of number of iterations on accuracy in the Bamboogle and DeepSearch benchmarks.}
    \label{fig:hyper-parameter}
    \vspace{-6pt}
\end{wrapfigure}

We investigate the impact of the optimization depth on system performance by varying the number of evolution iterations, as illustrated in Figure~\ref{fig:hyper-parameter}. Each iteration corresponds to a complete cycle of trajectory analysis, experience extraction, and FSM refinement. 
The results, based on the GPT-4o backbone, demonstrate a clear positive correlation between iterative optimization and task accuracy. Notably, on the highly complex DeepSearch benchmark, we observe a substantial 16\% performance gain as iterations increase, suggesting that intricate research tasks benefit significantly from continuous structural refinement.
In contrast, for the relatively straightforward Bamboogle benchmark, performance rapidly plateaus after just three iterations. This saturation indicates that while our self-evolution mechanism is highly effective, the optimal optimization depth is task-dependent. Practically, this suggests that EvoFSM can adaptively balance performance gains against computational overhead, investing more self-evolving steps only where the problem complexity demands it.

\paragraph{Cases Analysis and Discussion.}
The presented cases qualitatively validate the core hypothesis of EvoFSM: that robust self-evolution requires the distinct treatment of workflow topology and individual expertise.
As shown in Figure~\ref{fig:case_flow}, structural bottlenecks like infinite search loops typically withstand simple prompt adjustments and instead necessitate a topological intervention, such as \texttt{ADD\_STATE}, to fundamentally alter the reasoning path.
Conversely, Figure~\ref{fig:case_instruction} demonstrates that when the workflow is sound but the output lacks precision, targeted \texttt{REVISE\_INSTRUCTION} operations effectively sharpen node-specific skills without disrupting the global process.
Most importantly, Figure~\ref{fig:case_synergy} highlights the necessity of synergistic evolution for complex tasks. By simultaneously deploying a \texttt{Verifier} node to enforce source quality and refining the \texttt{Search Agent} to target legal terminologies, EvoFSM successfully transforms a generic retrieval pipeline into a rigorous legal analysis workflow.
Collectively, these examples illustrate that our structured, decoupled framework enables agents to precisely diagnose and repair specific failure modes. This capability stands in sharp contrast to unstructured rewriting approaches, which often struggle to isolate such orthogonal issues, leading to unstable or unpredictable evolution.

\section{Conclusion}
In this paper, we presented \textbf{EvoFSM}, a structured self-evolving framework designed to overcome the limitations of both static workflows and unconstrained self-evolving agents in deep research tasks. By formalizing the research process as a dynamic Finite State Machine, EvoFSM decouples the optimization space into macroscopic \textit{Flow} and microscopic \textit{Skill}. Unlike existing methods that rely on unstructured global rewriting, our approach empowers system to adapt through a set of precise atomic operations, ensuring that the self-evolving process remains targeted and controllable. Furthermore, the integration of a self-evolving memory mechanism allows the system to distill and transfer successful exploration strategies across tasks, facilitating continuous learning. Extensive experiments show that our EvoFSM significantly outperforms strong baselines on five multi-hop QA benchmarks while exhibiting superior generalization in interactive decision-making environments. We believe this paradigm shift—from black-box self-modification to structured, controllable evolution—provides a promising path toward reliable autonomous agents that can tackle complex, open-ended real-world problems.

\section*{Limitations}

Despite the promising performance of EvoFSM in automating deep research, three key limitations remain to be addressed in future work.
First, our framework currently relies entirely on off-the-shelf proprietary LLMs via prompt engineering and in-context learning. We do not perform any fine-tuning or specialized training on the underlying models. While this ensures broad accessibility, it inherently limits the system's efficiency and responsiveness, as general-purpose models may struggle to internalize complex FSM logic without explicit weight updates. Future iterations could benefit significantly from distilling these self-evolving capabilities into smaller, specialized agents.
Second, the reliability of the entire self-evolving process hinges on the Critic Mechanism. Since the system operates without external ground truth during deployment, it depends on the Critic (itself an LLM) to accurately diagnose failures. If the Critic hallucinates a successful verification or fails to detect subtle logical errors, the system may learn incorrect patterns or fail to evolve effectively. Developing more robust, verification-guided critics remains an open challenge.
Finally, as the agent continuously solves new tasks, the self-evolving memory faces a scalability issue. The experience pool currently grows indefinitely without a mechanism for consolidation or forgetting. Over time, this accumulation may lead to retrieval latency and the retrieval of redundant or outdated strategies. Implementing a long-term memory management system that can abstract, merge, or prune experiences is essential for sustaining performance in life-long learning scenarios.

\newpage

\bibliography{custom}

\clearpage
\onecolumn
\appendix

\section{Illustrative Examples}
\label{sec:appendix_cases}

\begin{center}
    \begin{tcolorbox}[colback=yellow!6!white, colframe=blue!50!green, title=\textbf{Case 1: Flow Evolution via \texttt{ADD\_STATE}}]
    \small
    \textbf{User Query:} "What are the specific environmental impacts of the continuous construction of the Three Gorges Dam recorded in 2023 reports?"

    \medskip
    \textbf{Initial FSM Execution (Failed):}
    \begin{itemize}
        \item \texttt{[Search Agent]} $\rightarrow$ Queries "Three Gorges Dam environmental impact 2023".
        \item \texttt{[Browse Agent]} $\rightarrow$ Reads generic Wikipedia pages.
        \item \texttt{[Search Agent]} $\rightarrow$ Queries again (Loop detected).
        \item \textit{Outcome:} The system enters a "Search-Browse" infinite loop, unable to find specific 2023 reports, leading to a timeout.
    \end{itemize}

    \textbf{Evolution (Flow Dimension):}
    \begin{itemize}
        \item \textit{Diagnosis:} The standard loop lacks a mechanism to verify the \textit{specificity} of the date in retrieved documents.
        \item \textit{Atomic Operation:} \texttt{ADD\_STATE(Verifier)}.
        \item \textit{New Topology:} \texttt{Search} $\rightarrow$ \texttt{Browse} $\rightarrow$ \textbf{\texttt{Verifier}} $\rightarrow$ \texttt{Synthesize}.
    \end{itemize}

    \textbf{Evolved FSM Execution (Success):}
    \begin{itemize}
        \item ... (Search \& Browse steps) ...
        \item \textbf{\texttt{[Verifier Agent]}} $\rightarrow$ "The retrieved documents are from 2020. We need specific 2023 data. Refining query to 'Three Gorges Dam annual report 2023 pdf'."
        \item \texttt{[Search Agent]} $\rightarrow$ Executes refined query.
        \item \textit{Outcome:} Successfully locates the 2023 annual environmental report and extracts specific data points.
    \end{itemize}
    \end{tcolorbox}
    \captionof{figure}{An example of \textit{Flow Evolution}. EvoFSM identifies a reasoning deadlock and structurally intervenes by injecting a new \texttt{Verifier} state, enabling the system to break out of a low-quality search loop.}
    \label{fig:case_flow}
\end{center}

\vspace{-0.5cm} 

\begin{center}
    \begin{tcolorbox}[colback=yellow!6!white, colframe=blue!50!green, title=\textbf{Case 2: Instruction Evolution via \texttt{REVISE\_INSTRUCTION}}]
    \small
    \textbf{User Query:} "Compare the battery energy density of the latest EV models from Tesla, BYD, and Nio launched in Q4 2023."

    \medskip
    \textbf{Initial FSM Execution (Suboptimal):}
    \begin{itemize}
        \item \texttt{[Search Agent]} $\rightarrow$ Finds comparison articles.
        \item \texttt{[Browse Agent]} $\rightarrow$ Summarizes content: "Tesla has high density, BYD uses Blade battery..."
        \item \textit{Outcome:} The answer is vague and qualitative, missing the specific "Wh/kg" numerical values required for a rigorous comparison.
    \end{itemize}

    \textbf{Evolution (Instruction Dimension):}
    \begin{itemize}
        \item \textit{Diagnosis:} The \texttt{Browse Agent} is summarizing too aggressively, losing critical quantitative details.
        \item \textit{Atomic Operation:} \texttt{REVISE\_INSTRUCTION(Browse Agent)}.
        \item \textit{Modification:} Append constraint: "Do not summarize numerical data. Extract exact values with units (e.g., Wh/kg) verbatim from the text."
    \end{itemize}

    \textbf{Evolved FSM Execution (Success):}
    \begin{itemize}
        \item \texttt{[Browse Agent] (Evolved)} $\rightarrow$ Extracts: "Tesla Model 3 Highland: 260 Wh/kg; BYD Seal: 150 Wh/kg..."
        \item \textit{Outcome:} A precise, data-driven comparison table is generated.
    \end{itemize}
    \end{tcolorbox}
    \captionof{figure}{An example of \textit{Instruction Evolution}. EvoFSM detects information loss in the \texttt{Browse} node and fine-tunes its system prompt without altering the overall workflow, achieving higher precision.}
    \label{fig:case_instruction}
\end{center}

\vspace{1cm}

\begin{center}
    \begin{tcolorbox}[colback=yellow!6!white, colframe=blue!50!green, title=\textbf{Case 3: Synergistic Evolution (Flow + Skill)}]
    \small
    \textbf{User Query:} "Analyze how the EU AI Act (late 2023 draft) regulates open-source foundation models compared to proprietary ones, citing specific Articles."

    \medskip
    \textbf{Phase 1: Initial Failure (Static FSM)}
    \begin{itemize}
        \item \textbf{Execution:} \texttt{Search Agent} queries "EU AI Act open source". \texttt{Browse Agent} summarizes TechCrunch articles.
        \item \textbf{Failure Mode:} The system produces a vague summary ("It has exemptions") without citing specific articles, as it relies on secondary news sources.
        \item \textbf{Diagnosis:} 
            1. \textit{Flow Deficit}: The workflow lacks a verification step to distinguish official legal texts from news.
            2. \textit{Skill Deficit}: The \texttt{Search Agent} uses generic keywords instead of legal terminology.
    \end{itemize}

    \textbf{Phase 2: Dual-Dimension Evolution}
    \begin{itemize}
        \item \textbf{Flow Action ($\mathcal{O}_{flow}$):} \texttt{ADD\_STATE(Legal\_Verifier)}. 
        \textit{Purpose:} To explicitly filter retrieved documents for official legislative formatting (e.g., "Recital", "Article").
        \item \textbf{Skill Action ($\mathcal{O}_{skill}$):} \texttt{REVISE\_INSTRUCTION(Search Agent)}. 
        \textit{Refinement:} "Constraint: Do not search for news. Construct queries targeting specific legal sections (e.g., 'EU AI Act Article 53 exemption')."
    \end{itemize}

    \textbf{Phase 3: Evolved Success}
    \begin{itemize}
        \item \textbf{New Workflow:} \texttt{Search (Legal-Focused)} $\rightarrow$ \texttt{Browse} $\rightarrow$ \textbf{\texttt{Legal\_Verifier}} $\rightarrow$ \texttt{Synthesize}.
        \item \textbf{Execution:} 
            1. Search Agent executes a precise query.
            2. Browse Agent reads the official PDF.
            3. \texttt{Legal\_Verifier} confirms: "Found Recital 60i and Article 53(2)."
        \item \textbf{Outcome:} Synthesizes a legally accurate comparison citing the exact clauses.
    \end{itemize}
    \end{tcolorbox}
    \captionof{figure}{An example of \textit{Synergistic Evolution}. EvoFSM simultaneously reconfigures the collaboration topology (adding a Verifier) and sharpens the individual search expertise. This dual optimization enables the agent to pivot from superficial news summarization to rigorous legal analysis.}
    \label{fig:case_synergy}
\end{center}

\end{document}